\documentclass[letterpaper]{article}
\usepackage{aaai}
\usepackage{times}
\usepackage{helvet}
\usepackage{courier}
\usepackage{longtable,multirow}
\usepackage[utf8]{inputenc} 
\usepackage[T1]{fontenc}    
\usepackage{url}            
\usepackage{booktabs}       
\usepackage{amsfonts}       
\usepackage{nicefrac}       
\usepackage{microtype}      
\usepackage{amssymb,amsmath,amsfonts}
\usepackage{color}
\usepackage{graphicx,xspace,soul}
\usepackage{epsfig,subfigure}
\usepackage{amsthm}
\usepackage{soul}

\def\0{{\mathbf 0}}
\def\1{{\mathbf 1}}

\def\b{{\mathbf b}}

\def\e{{\mathbf e}}
\def\f{{\mathbf f}}

\def\v{{\mathbf v}}

\def\x{{\mathbf x}}
\def\y{{\mathbf y}}
\def\z{{\mathbf z}}

\def\A{{\mathbf A}}
\def\B{{\mathbf B}}
\def\C{{\mathbf C}}
\def\D{{\mathbf D}}
\def\E{{\mathbf E}}
\def\F{{\mathbf F}}

\def\F{{\mathbf F}}
\def\H{{\mathbf H}}

\def\L{{\mathbf L}}
\def\M{{\mathbf M}}

\def\Q{{\mathbf Q}}

\def\S{{\mathbf S}}
\def\T{{\mathbf T}}

\def\W{{\mathbf W}}
\def\X{{\mathbf X}}

\def\ie{{\textit{i.e.}}}

\def\cE{{\mathcal E}}

\def\cG{{\mathcal G}}

\def\cL{{\mathcal L}}

\def\cO{{\mathcal O}}

\def\cS{{\mathcal S}}

\def\cV{{\mathcal V}}

\begin{document}
%
\title{Unfolding Projection-free SDP Relaxation of Binary Graph Classifier via \\
GDPA Linearization}
\author{Cheng Yang\,$^\dagger$, Gene Cheung\,$^\ddagger$, Wai-tian Tan\,$^\mathsection$, Guangtao Zhai\,$^\dagger$\\
$^\dagger$\,Shanghai Jiaotong University, Shanghai, China $^\ddagger$\,York University, Toronto, Canada $^\mathsection$\,Cisco Systems, San José, CA\\
}
\maketitle
\begin{abstract}
\begin{quote}
Algorithm unfolding creates an interpretable and parsimonious neural network architecture by implementing each iteration of a model-based algorithm as a neural layer. 
However, unfolding a proximal splitting algorithm with a positive semi-definite (PSD) cone projection operator per iteration is expensive, due to the required full matrix eigen-decomposition.
In this paper, leveraging a recent linear algebraic theorem called Gershgorin disc perfect alignment (GDPA), we unroll a projection-free algorithm for semi-definite programming relaxation (SDR) of a binary graph classifier, where the PSD cone constraint is replaced by a set of ``tightest possible'' linear constraints per iteration.
As a result, each iteration only requires computing a linear program (LP) and one extreme eigenvector.
Inside the unrolled network, we optimize parameters via stochastic gradient descent (SGD) that determine graph edge weights in two ways: i) a metric matrix that computes feature distances, and ii) a sparse weight matrix computed via local linear embedding (LLE).
Experimental results show that our unrolled network outperformed pure model-based graph classifiers, and achieved comparable performance to pure data-driven networks but using far fewer parameters. 
\end{quote}
\end{abstract}

\section{INTRODUCTION}
\label{sec:intro}
While generic and powerful \textit{deep neural networks} (DNN) \cite{dlnature} can achieve state-of-the-art performance using large labelled datasets for many data-fitting problems such as image restoration and classification \cite{Zhang_2017_CVPR,NIPS2012_c399862d}, they operate as ``black boxes'' that are difficult to explain. 
To build an interpretable system targeting a specific problem instead, \textit{algorithm unfolding} \cite{9363511} takes a model-based iterative algorithm, implements (unrolls) each iteration as a neural layer, and stacks them in sequence to compose a network architecture. 
As a pioneering example, LISTA \cite{10.5555/3104322.3104374} implemented each iteration of a sparse coding algorithm called ISTA \cite{Beck2009AFI}---composed of a gradient descent step and a soft thresholding step---as linear and ReLU operators in a neural layer. 
By optimizing two matrix parameters in the linear operator per layer end-to-end via \textit{stochastic gradient descent} (SGD) \cite{bottou-98x}, LISTA converged faster and had better performance. 
This means that the required iteration / neural layer count was comparatively small, resulting in a parsimonious architecture with few learned network parameters. 

However, algorithm unfolding is difficult if the iterative algorithm performs \textit{proximal splitting} \cite{boyd11} with a \textit{positive semi-definite} (PSD) cone projection operator per iteration; PSD cone projection is common in algorithms solving a \textit{semi-definite programming} (SDP) problem with a PSD cone constraint \cite{gartner12}.
A PSD cone projection for a matrix variable $\H$ requires full matrix eigen-decomposition on $\H$ with complexity $\cO(N^3)$. 
Not only is the computation cost of the projection in a neural layer expensive, optimizing network parameters through the projection operator via SGD is difficult.
\begin{figure}[t]
\begin{center}
\includegraphics[trim={0cm 12.1cm 14cm 0cm},clip,width=0.95\linewidth]{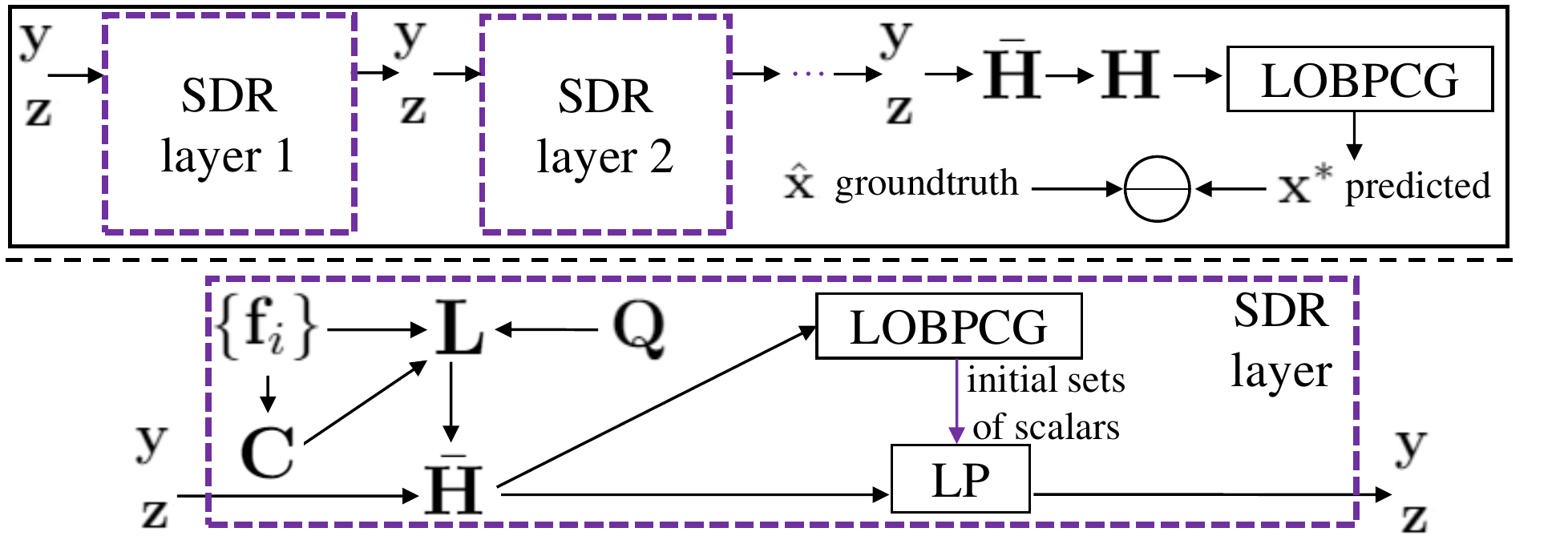}
\end{center}
\vspace{-0.15in}
\caption{Overview of our SDP relaxation (SDR) network. 
Upper: SDR network composed of stacked SDR layers.
Lower: The architecture of a single SDR layer.
The lower triangluar matrix $\Q$ and local linear embedding (LLE) weight matrix $\C$ are used to define the conically combined Laplacian matrix $\L$, which is then used together with the SDP dual variables $\y$ and $\z$ to define the matrix $\bar{\H}$.
Next, $\bar{\H}$ is passed to differentiable LOBPCG solver and LP solver to update $\y$ and $\z$.
The network is end-to-end trained with an MSE loss of the predicted labels.} 
\label{fig:gdpaunroll}
\end{figure}

In this paper, using binary graph classifier \cite{zhou03,belkin04,guillory09,5447068} as an illustrative application, we demonstrate how PSD cone projection can be entirely circumvented for an SDP problem, facilitating algorithm unfolding and end-to-end optimization of network parameters without sacrificing performance. 
Specifically, we first replace the PSD cone constraint in the original \textit{semi-definite programming relaxation} (SDR) \cite{li08} of the NP-hard graph classifier problem with ``tightest possible'' linear constraints per iteration, thanks to a recent linear algebraic theorem called \textit{Gershgorin disc perfect alignment} (GDPA) \cite{yang2021signed}.
Together with the linear objective, each iteration computes only a \textit{linear program} (LP) \cite{vanderbei21} and one extreme eigenvector (computable in $\cO(N)$ using LOBPCG \cite{Knyazev01}).

We next unroll the now projection-free iterative algorithm into an interpretable network, and optimize parameters that determine graph edge weights per neural layer via SGD in two ways.
First, assuming edge weight $w_{i,j}$ is inversely proportional to \textit{feature distance} $d_{i,j}$ between nodes $i$ and $j$ endowed with feature vectors $\f_i$ and $\f_j$ respectively, we optimize a PSD metric matrix $\M$ via Cholesky factorization \cite{GoluVanl96} $\M = \Q \Q^{\top}$ that computes \textit{Mahalanobis distance} \cite{mahalanobis1936} as $d_{i,j} = (\f_i - \f_j)^{\top} \M (\f_i - \f_j)$.
Second, we initialize a non-negative symmetric weight matrix via \textit{local linear embedding} (LLE) \cite{Roweis2000NonlinearDR,ghojogh2020locally} given feature vectors $\f_i$'s, which we subsequently fine-tune per layer in a semi-supervised manner.
We employ a conic combination of the two resulting graph Laplacian matrices for classification in each layer. 
An illustration of the unrolled network is shown in Fig.\;\ref{fig:gdpaunroll}. 

We believe this methodology of replacing the PSD cone constraint by linear constraints per iteration---leading to an iterative algorithm amenable to algorithm unfolding---can be more generally applied to a broad class of SDP problems with PSD cone constraints \cite{gartner12}.
For binary graph classifiers, experimental results show that our interpretable unrolled network substantially outperformed pure model-based classifiers \cite{yang2021projectionfree}, and achieved comparable performance as pure data-driven networks \cite{dlnature} but using noticeably fewer parameters.

\section{RELATED WORK}
\label{sec:related}
Algorithm unfolding is one of many classes of approaches in \textit{model-based deep learning} \cite{shlezinger2021modelbased}, and has been shown effective in creating interpretable network architectures for a range of data-fitting problems \cite{9363511}. 
We focus on unfolding of iterative algorithms involving PSD cone projection  \cite{odonoghue16} that are common when addressing SDR of NP-hard \textit{quadratically constrained quadratic programming} (QCQP) problems \cite{5447068}, of which binary graph classifier is a special case. 

Graph-based classification was first studied two decades ago \cite{zhou03,belkin04,guillory09}.
An interior point method tailored for the slightly more general \textit{binary quadratic problem}\footnote{BQP objective takes a quadratic form $\x^{\top} \Q \x$, but $\Q$ is not required to be a Laplacian matrix to a similarity graph.} (BQP) has complexity $\cO(N^{3.5} \log (1/\epsilon))$, where $\epsilon$ is the tolerable error \cite{helmberg96}.
Replacing PSD cone constraint $\M \succeq 0$ with a factorization $\M = \X \X^{\top}$ was proposed \cite{shah16}, but it resulted in a non-convex optimization for $\X$ that was solved locally via alternating minimization, where in each iteration a matrix inverse of worst-case complexity $\cO(N^3)$ was required.  
More recent first-order methods such as \cite{odonoghue16} used ADMM \cite{boyd11}, but still requires expensive PSD cone projection per iteration.
In contrast, leveraging GDPA theory \cite{yang2021signed}, our algorithm is entirely projection-free.

GDPA theory was developed for \textit{metric learning} \cite{moutafis17} to optimize a PD \textit{metric matrix} $\M$, given a convex and differentiable objective $Q(\M)$, in a Frank-Wolfe optimization framework \cite{pmlr-v28-jaggi13}. 
This paper leverages GDPA \cite{yang2021signed} in an entirely different direction for unfolding of a projection-free graph classifier learning algorithm. 

\section{PRELIMINARIES}
\label{sec:prelim}
\subsection{Graph Definitions}

A graph is defined as $\cG(\cV,\cE,\W)$, with node set $\cV = \{1 \ldots, N\}$, and edge set  $\cE = \{ (i,j)\}$, where $(i,j)$ means nodes $i$ and $j$ are connected with weight $w_{i,j} \in \mathbb{R}$.
A node $i$ may have a self-loop of weights $u_i \in \mathbb{R}$. 
Denote by $\W$ the \textit{adjacency matrix}, where $W_{i,j} = w_{i,j}$ and $W_{i,i} = u_i$. 
We assume that edges are undirected, and $\W$ is symmetric.
Define next the diagonal \textit{degree matrix} $\D$, where $D_{i,i} = \sum_{j} W_{i,j}$. 
The \textit{combinatorial graph Laplacian matrix} \cite{ortega18ieee} is then defined as $\L \triangleq \D - \W$.
To account for self-loops, the \textit{generalized graph Laplacian matrix}  is defined as $\cL \triangleq \D - \W + \text{diag}(\W)$.
Note that any real symmetric matrix can be interpreted as a generalized graph Laplacian matrix. 

The \textit{graph Laplacian regularizer} (GLR) \cite{pang17} that quantifies smoothness of signal $\x \in \mathbb{R}^N$ w.r.t. graph specified by $\cL$ is
\begin{align}
\x^{\top} \cL \x = \sum_{(i,j) \in \cE} w_{i,j} (x_i - x_j)^2 + \sum_{i \in \cV} u_i x_i^2 .
\label{eq:glr}
\end{align}
GLR is also the objective of our graph-based classification problem.

\subsection{GDPA Linearization}

\begin{figure}[t]
\begin{small}
\begin{align*}
\hspace{-0.0in}\mathbf{M}&=\left[ \begin{array}{ccc}
2 & -2 & -1\\
-2 & 5 & -2\\
-1 & -2 & 4\\
\end{array} \right]\\
\hspace{-0.0in}\mathbf{SMS}^{-1}&=
\left[ \begin{array}{ccc}
2 & -1.301 & -0.5912\\
-3.0746 & 5 & -1.8176\\
-1.6915 & -2.2007 & 4\\
\end{array} \right]
\end{align*} 
\end{small}
\vspace{-0.3in}
\begin{center}
\hspace{0.1in}
\includegraphics[width=0.99\linewidth]{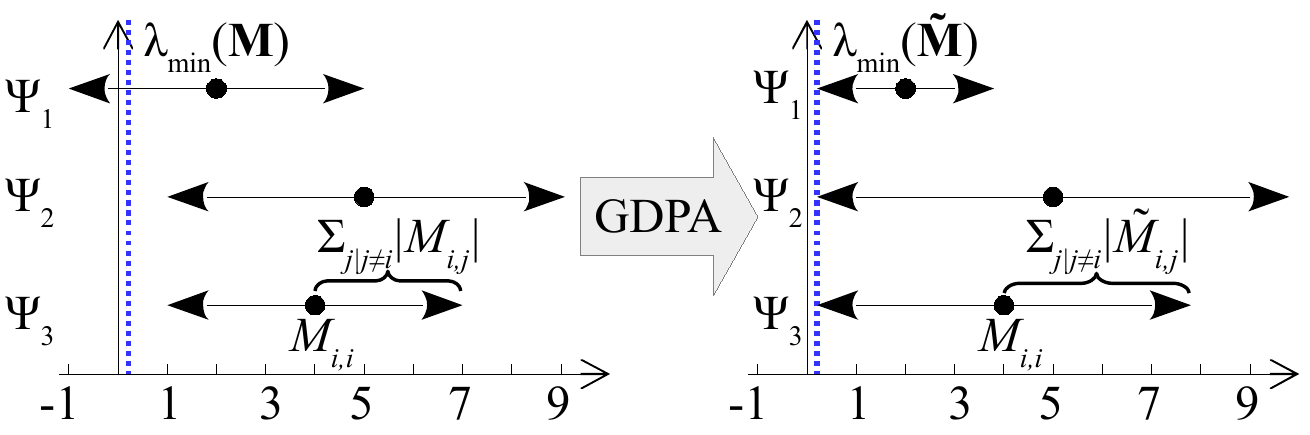}
\end{center}
\vspace{-0.15in}
\caption{\small Example of a PD matrix $\M$ and its similarity transform $\tilde{\M} = \S \M \S^{-1}$, and their respective Gershgorin discs $\Psi_i$. 
Gershgorin disc left-ends of $\tilde{\M}$ are aligned at $\lambda_{\min}(\M) = 0.1078$.}
\label{fig:GDPA}
\end{figure}

To ensure matrix variable $\M$ is PSD without eigen-decomposition, we leverage GDPA \cite{yang2021signed}.
Given a real symmetric matrix, we define a \textit{Gershgorin disc} $\Psi_i$ corresponding to row $i$ of $\M$ with center $c_i(\M) \triangleq M_{i,i}$ and radius $r_i(\M) \triangleq \sum_{j \neq i} |M_{i,j}|$.
By \textit{Gershgorin Circle Theorem} (GCT) \cite{gct}, the smallest real eigenvalue $\lambda_{\min}(\M)$ of $\M$ is lower-bounded by the smallest disc left-end $\lambda^-_{\min}(\M)$, \ie,
\begin{align}
\lambda^-_{\min}(\M) \triangleq \min_i c_i(\M) - r_i(\M) \leq \lambda_{\min}(\M) .
\label{eq:GCT_bound}
\end{align}
Thus, to ensure $\M \succeq 0$, one can impose the sufficient condition $\lambda^-_{\min}(\M) \geq 0$, or equivalently 
\begin{align}
c_i(\M) - r_i(\M) \geq 0, ~~~\forall i.
\end{align}
However, GCT lower bound $\lambda^-_{\min}(\M)$ tends to be loose.
As an example, consider the \textit{positive definite} (PD) matrix $\M$ in Fig.\;\ref{fig:GDPA} with $\lambda_{\min}(\M) = 0.1078$. 
The first disc left-end is $c_1(\M) - r_1(\M) = 2 - 3 = -1$, and $\lambda^-_{\min}(\M) < 0$.

GDPA provides a theoretical foundation to tighten the GCT lower bound.
Specifically, GDPA states that given a generalized graph Laplacian matrix $\M$ corresponding to a ``balanced'' signed graph\footnote{A balanced graph has no cycles of odd number of negative edges. By the Cartwright-Harary Theorem, a graph is balanced iff nodes can be colored into red/blue, so that each positive/negative edge connects nodes of the same/different colors.} $\cG$ \cite{cartwright56}, one can perform a \textit{similarity transform}\footnote{A similarity transform $\B = \S \A \S^{-1}$ and the original matrix $\A$ share the same set of eigenvalues \cite{gct}.}, 
$\tilde{\M} = \S \M \S^{-1}$, where $\S = \text{diag}(v_1^{-1}, \ldots, v_N^{-1})$ and $\v$ is the first eigenvector of $\M$, 
such that all the disc left-ends of $\tilde{\M}$ are exactly aligned at $\lambda_{\min}(\M) = \lambda_{\min}(\tilde{\M})$. 
This means that transformed $\tilde{\M}$ satisfies $\lambda^-_{\min}(\tilde{\M}) = \lambda_{\min}(\tilde{\M})$; \ie, \textit{the GCT lower bound is the tightest possible after an appropriate similarity transform}. 
Continuing our example, similarity transform $\tilde{\M} = \S \M \S^{-1}$ of $\M$ has all its disc left-ends exactly aligned at $\lambda_{\min}(\M) = \lambda_{\min}(\tilde{\M}) = 0.1078$. 

Leveraging GDPA, \cite{yang2021signed} developed a fast metric learning algorithm, in which the PSD cone constraint $\M \succeq 0$ is replaced by linear constraints $\lambda^-_{\min}(\S_t \M \S_t^{-1}) \geq 0$ per iteration, where $\S_t = \text{diag}(v_1^{-1}, \ldots, v_N^{-1})$ and $\v$ is the first eigenvector of previous solution $\M_{t-1}$. 
Assuming that the algorithm always seeks solutions $\M$ in the space of graph Laplacian matrices of balanced graphs, this means previous PSD solution $\M_{t-1}$ remains feasible at iteration $t$, since by GDPA $\lambda^-_{\min}(\S_t \M_{t-1} \S_{t}^{-1}) = \lambda_{\min}(\M_{t-1}) \geq 0$.
Together with a convex and differentiable objective, the optimization can thus be solved efficiently in each iteration using the projection-free Frank-Wolfe procedure \cite{pmlr-v28-jaggi13}. 
This process of computing the first eigenvector $\v$ of a previous PSD solution $\M_{t-1}$ to establish linear constraints $\lambda^-_{\min}(\S_t \M \S_t^{-1}) \geq 0$ in the next iteration, replacing the PSD cone constraint $\M \succeq 0$, is called \textit{GDPA linearization}.

\section{GRAPH CLASSIFIER LEARNING}
\label{sec:formulate}
We first formulate the binary graph classifier learning problem and relax it to an SDP problem.
We then present its SDP dual with dual variable matrix $\H$.
Finally, we augment the SDP dual with variable  $\bar{\H}$, which is a graph Laplcian to a balanced graph, amenable to GDPA linearization.

\subsection{SDP Primal}
\label{subsec:SDP_primal}

Given a PSD graph Laplacian matrix $\L \in \mathbb{R}^{N \times N}$ of a \textit{positive} similarity graph $\cG^o$ (\ie, $w_{i,j} \geq 0, \forall (i,j) \in \cE$), we formulate a graph-based binary classification problem as
\begin{align}
\min_{\x} \x^{\top} \L \x, 
~~~~ \mbox{s.t.}~ \left\{
\begin{array}{l}
x_i^2 = 1, ~\forall i \in \{1, \ldots, N\} \\
x_i = \hat{x}_i, ~\forall i \in \{1, \ldots, M\}
\end{array} \right. .
\label{eq:binaryClass}
\end{align}
where $\{\hat{x}_i\}_{i=1}^M$ are the $M$ known binary labels. 
The quadratic objective in \eqref{eq:binaryClass} is a GLR \eqref{eq:glr}, promoting a label solution $\x$ that is smooth w.r.t. graph $\cG^o$ specified by $\L$. 
The first constraint ensures $x_i$ is binary, \ie, $x_i \in \{-1, 1\}$. 
The second constraint ensures that entries in $\x$ agree with known labels $\{\hat{x}_i\}_{i=1}^M$. 

\begin{figure}[t]
\begin{center}
\includegraphics[width=0.95\linewidth]{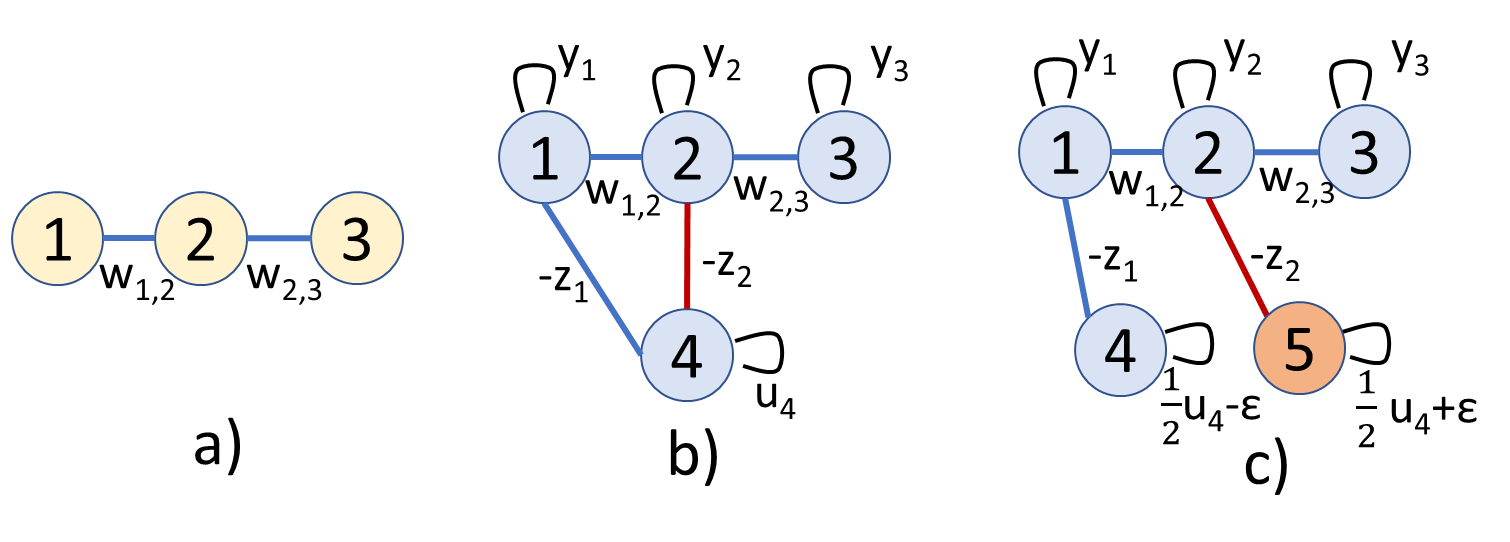}
\end{center}
\vspace{-0.3in}
\caption{\small (a) An example 3-node line graph. (b) Unbalanced graph corresponding to solution $\H$ to SDR dual \eqref{eq:SDP_dual} interpreted as graph Laplacian matrix. (c) Balanced graph corresponding to solution $\bar{\H}$ to modified SDR dual \eqref{eq:SDP_dual2} interpreted as Laplacian. 
Positive / negative edges are colored in blue / red. Self-loop weight $u_4$ in (b) for node $4$ is $u_4 = y_4 + z_1 + z_2$.}
\label{fig:3-node-graph}
\end{figure}

As an example, consider a 3-node line graph shown in Fig.\;\ref{fig:3-node-graph}(a), where edges $(1,2)$ and $(2,3)$ have weights $w_{1,2}$ and $w_{2,3}$, respectively. 
The corresponding adjacency and graph Laplacian matrices, $\W$ and $\L$, are:
\begin{scriptsize}
\begin{align*}
\W = \left[ \begin{array}{ccc}
0 & w_{1,2} & 0 \\
w_{1,2} & 0 & w_{2,3} \\
0 & w_{2,3} & 0
\end{array} \right],  
\L = \left[ \begin{array}{ccc}
d_1 & -w_{1,2} & 0 \\
-w_{1,2} & d_2 & -w_{2,3} \\
0 & -w_{2,3} & d_3
\end{array} \right]
\end{align*}
\end{scriptsize}\noindent
where $d_i = \sum_{j} w_{i,j}$ is the \textit{degree} of node $i$. 
Suppose known labels are $\hat{x}_1 = 1$ and $\hat{x}_2 = -1$. 

\eqref{eq:binaryClass} is NP-hard due to the binary constraint. 
One can define a corresponding SDR problem as follows.
Define first matrix $\X = \x \x^{\top}$, then $\M = [\X ~~ \x; ~\x^{\top} ~~ 1]$.
$\M$ is PSD because: 
i) block $[1]$ is PSD, and ii) the \textit{Schur complement} of block $[1]$ of $\M$ is $\X - \x \x^{\top} = \0$, which is also PSD.
Thus, $\X = \x \x^{\top}$ (\ie, $\text{rank}(\X) = 1$) implies $\M \succeq 0$.
$\X = \x \x^{\top}$ and $X_{ii} = 1, \forall i$ together imply $x_i^2 = 1, \forall i$. 
To convexify the problem, we drop the non-convex rank constraint and write the SDR as
\begin{align}
\min_{\x,\X} \text{Tr}(\L \X) ~~\mbox{s.t.}~~
\left\{ \begin{array}{l}
X_{ii} = 1, i \in \{1, \ldots, N\} \\
\M \triangleq \left[ \begin{array}{cc}
\X & \x \\
\x^{\top} & 1
\end{array} \right] \succeq 0 \\
x_i = \hat{x}_i, ~~ i \in \{1, \ldots, M\}
\end{array} \right. 
\label{eq:primal}
\end{align}
where $\text{Tr}(\x^{\top} \L \x) = \text{Tr}(\L \x \x^{\top}) = \text{Tr}(\L \X)$. 
Because \eqref{eq:primal} has linear objective and constraints with an additional PSD cone constraint, $\M \succeq 0$, it is an SDP problem. 
We call \eqref{eq:primal} the \textit{SDR primal}.

Unfortunately, the solution $\M$ to \eqref{eq:primal} is not a graph Laplacian matrix to a balanced graph, and hence GDPA linearization cannot be applied.
Thus, we next investigate its SDP dual instead.

\subsection{SDP Dual with Balanced Graph Laplacian}
\label{subsec:SDP_dual}

Following standard SDP duality theory  \cite{gartner12}, we write the corresponding dual problem as follows.
We first define
\begin{align}
\A_i = \text{diag}(\e_{N+1}(i)),
~~~~~
\B_i = \left[ \begin{array}{cc}
\0_{N \times N} & \e_N(i) \\
\e^{\top}_N(i) & 0
\end{array}
\right] 
\label{eq:AB}
\end{align}
where $\e_N(i) \in \{0, 1\}^{N}$ is a length-$N$ binary \textit{canonical vector} with a single non-zero entry equals to $1$ at the $i$-th entry, $\0_{N \times N}$ is a $N$-by-$N$ matrix of zeros, and $\text{diag}(\v)$ is a diagonal matrix with diagonal entries equal to $\v$.

Next, we put $M$ known binary labels $\{\hat{x}_i\}_{i=1}^M$ into a vector $\b \in \mathbb{R}^{M}$ of length $M$; specifically, we define
\begin{align}
b_i = 2 \hat{x}_i, 
~~~ \forall i \in \{1, \ldots, M\} .
\end{align}
We are now ready to write the SDR dual of \eqref{eq:primal} as 
\begin{footnotesize}
\begin{align}
\min_{\y, \z} & ~~ \1^{\top}_{N+1} \y + \b^{\top} \z, 
\label{eq:SDP_dual} \\
\mbox{s.t.} & ~~
\H \triangleq \sum_{i=1}^{N+1} y_i \A_i +
\sum_{i=1}^M z_i \B_i + \L_{N+1} \succeq 0
\nonumber 
\end{align}
\end{footnotesize}\noindent
where $\1_N$ is an all-one vector of length $N$, and $\L_{N+1} \triangleq [\L ~\0_{N \times 1}; \0_{1 \times N} ~0]$.
Variables to the dual \eqref{eq:SDP_dual} are $\y \in \mathbb{R}^{N+1}$ and $\z \in \mathbb{R}^M$. 

Given the minimization objective, when $b_i < 0$, the corresponding $z_i$ must be $\geq 0$, since $z_i < 0$ would make $\H$ harder to be PSD (a larger Gershgorin disc radius) while worsening the objective.
Similarly, for $b_i > 0$, $z_i \leq 0$.
Thus, the signs of $z_i$'s are known beforehand.
Without loss of generality, we assume $z_i \leq 0, \forall i \in \{1, \ldots, M_1\}$ and $z_i \geq 0, \forall i \in \{M_1+1, \ldots, M\}$ in the sequel. 

Continuing our earlier 3-node graph example, solution $\H$ to the SDP dual \eqref{eq:SDP_dual} is
\begin{align}
\H = \left[ \begin{array}{cccc}
y_1 + d_1 & -w_{1,2} & 0 & z_1 \\
-w_{1,2} & y_2 + d_2 & -w_{2,3} & z_2 \\
0 & -w_{2,3} & y_3 + d_3 & 0 \\
z_1 & z_2 & 0 & y_4
\end{array} \right] .
\end{align}
The signed graph $\cG$ corresponding  to $\H$---interpreted as a generalized graph Laplacian matrix---is shown in Fig.\;\ref{fig:3-node-graph}(b). 
We see that the first three nodes correspond to the three nodes in Laplacian $\L$ with added self-loops of weights $y_i$'s.
The last node has $M=2$ edges with weights $-z_1$ and $-z_2$ to the first two nodes.
Because of the edges from the last node have different signs to the first $N$ nodes, $\cG$ is not balanced.

\subsection{Reformulating the SDP Dual}
\label{subsec:reformulate}

We construct a balanced graph $\bar{\cG}$ as an approximation to the imbalanced $\cG$.
This is done by splitting node $N+1$ in $\cG$ into two in $\bar{\cG}$, dividing positive and negative edges between them, as shown in Fig.\;\ref{fig:3-node-graph}. This results in $N+2$ nodes for $\bar{\cG}$.
The specific graph construction for $\bar{\cG}$ procedure is:
\begin{enumerate}
\item Construct first $N$ nodes with the same edges as $\cG$. 
\item Construct node $N+1$ with positive edges $\{-z_i\}_{i=1}^{M_1}$ and node $N+2$ with negative edges $\{-z_i\}_{i=M_1+1}^M$ to the first $N$ nodes in $\cG$.
\item Add self-loops for node $N+1$ and $N+2$ with respective weights $\bar{u}_{N+1} = u_{N+1}/2 - \epsilon$ and $\bar{u}_{N+2} = u_{N+1}/2 + \epsilon$, where $\epsilon \in \mathbb{R}$ is a parameter.  
\end{enumerate}

Denote by $\bar{\H} \in \mathbb{R}^{(N+2) \times (N+2)}$ the generalized graph Laplacian matrix to augmented graph $\bar{\cG}$. 
Continuing our example, Fig.\;\ref{fig:3-node-graph}(c) shows graph $\bar{\cG}$.
Corresponding $\bar{\H}$ is
\begin{footnotesize}
\begin{align*}
\bar{\H} = \left[ \begin{array}{ccccc}
y_1 + d_1 & -w_{1,2} & 0 & z_1 & 0 \\
-w_{1,2} & y_2 + d_2 & - w_{2,3} & 0 & z_2 \\
0 & -w_{2,3} & y_3 + d_3 & 0 & 0 \\
z_1 & 0 & 0 & \bar{u}_4 + z_1 & 0 \\
0 & z_2 & 0 & 0 & \bar{u}_5 + z_2
\end{array} \right] .
\end{align*}
\end{footnotesize}\noindent 
where $\bar{u}_4 = u_4/2 - \epsilon$, $\bar{u}_5 = u_4/2 + \epsilon$, and $u_4 = y_4 + z_1 + z_2$. 
Spectrally, $\bar{\H}$ and $\H$ are related; $\lambda_{\min}(\bar{\H}) \leq \lambda_{\min}(\H)$. 
See \cite{yang2021projectionfree} for a proof. 

We reformulate the SDP dual \eqref{eq:SDP_dual} by keeping the same objective but imposing PSD cone constraint on $\bar{\H}$ instead, which implies a PSD $\H$.
Define $\A'_i$, $\B'_i$ and $\B''_i$ similarly to \eqref{eq:AB} but for a larger $(N+2)$-by-$(N+2)$ matrix; \ie, $\A'_i = \text{diag}(\e_{N+2}(i))$, $\B'_i = [\B_i ~\0_{N+1}; \0_{N+1}^{\top} ~0]$, and $\B''_i = [\0_{(N+1) \times (N+1)} ~\e_{N+1}(i); \e^{\top}_{N+1}(i) ~0]$.
The reformulated SDR dual is
\begin{align}
\min_{\y, \z} &~~ \1^{\top}_{N+1} \y + \b^{\top} \z, 
\label{eq:SDP_dual2} \\
\mbox{s.t.} &~~
\bar{\H} \triangleq \sum_{i=1}^{N} y_i \A'_i +
\kappa_{N+1} \A'_{N+1} + 
\kappa_{N+2} \A'_{N+2} 
\nonumber \\
&~~~~~~~~~+\sum_{i=1}^{M_1} z_i \B'_i 
+ \sum_{i=M_1+1}^{M} z_i \B''_i 
- \L \succeq 0
\nonumber
\end{align}
where $\kappa_{N+1} = \frac{u_{N+1}}{2} - \sum_{i=1}^{M_1} z_i - \epsilon$ and $\kappa_{N+2} =  \frac{u_{N+1}}{2} - \sum_{i=M_1+1}^M z_i + \epsilon$. 

Given $\bar{\H}$ is now a Laplacian to a balanced graph, GDPA linearization can be applied to solve \eqref{eq:SDP_dual2} efficiently.
Specifically, in each iteration $t$, the first eigenvector $\v$ of previous solution $\bar{\H}_{t-1}$ is computed using LOBPCG to define matrix $\S_t = \text{diag}(v_1^{-1}, \ldots)$. 
$\S_t$ is then used to define linear constraints $\lambda^-_{\min}(\S_t \bar{\H} \S_t^{-1}) \geq 0$, replacing $\bar{\H} \succeq 0$ in \eqref{eq:SDP_dual2}. 
This results in a LP, efficiently solvable using a state-of-the-art LP solver such as Simplex or interior point \cite{vanderbei21}.
The algorithm is run iteratively until convergence. 


\section{OPTIMIZING GRAPH PARAMETERS}
\label{sec:dist}
After unrolling the iterative algorithm described above to solve \eqref{eq:SDP_dual2} into a neural network architecture as shown in Fig.\;\ref{fig:gdpaunroll}, we discuss next how to optimize parameters in each layer end-to-end via SGD for optimal performance. 
Specifically, we consider two methods---Mahalanobis distance learning and local linear embedding---to optimize graph edge weights, so that the most appropriate graph can be employed for classification in each layer.

\subsection{Mahalanobis Distance Learning}

We assume that edge weight $w_{i,j}$ between nodes $i$ and $j$ is inversely proportional to \textit{feature distance} $d_{i,j}$, computed using a Gaussian kernel, \ie, 
\begin{align}
w_{i,j} = \exp \left( - \frac{d_{i,j}}{\sigma_d^2} \right) .
\label{eq:edgeWeight}
\end{align}
Using an exponential kernel for $d_{i,j} \in [0, \infty)$ means $w_{i,j} \in (0, 1]$, 
which ensures a positive graph as required in \eqref{eq:binaryClass}. 

\begin{figure}
\begin{center}
\includegraphics[trim={0cm 0cm 0cm 0cm},clip,width=0.95\linewidth]{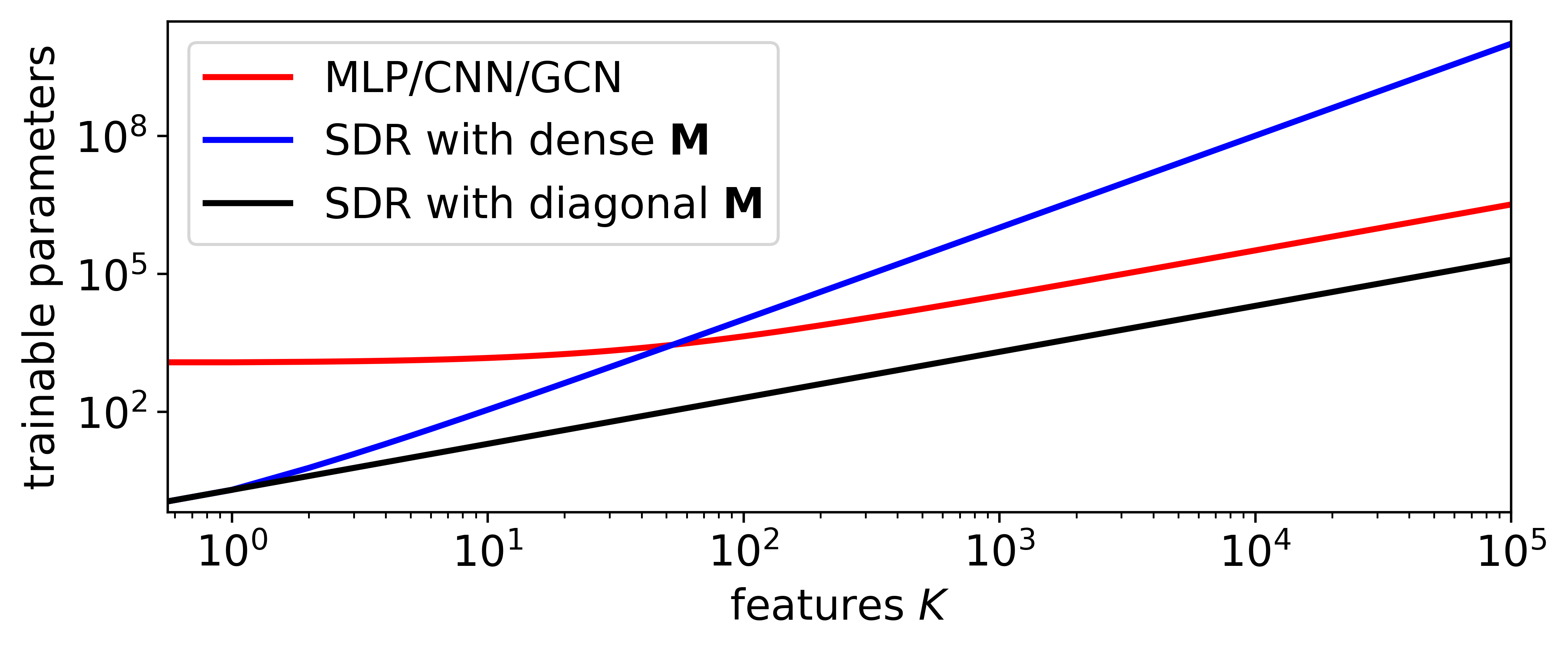}
\end{center}
\vspace{-0.25in}
\caption{Trainable parameters when the number of layers $P=2$ and number of neurons $M=32$ in a dense layer of a black-box network.
Our SDR can have trainable parameters close to $\mathcal{O}(K)$ (\textit{i.e.}, a line that leans towards the black line) with a sparsity pattern constraint.}
\label{fig:parameter_plot}
\end{figure}

We optimize feature distance $d_{i,j}$ in each neural layer as follows.
Assuming each node $i$ is endowed with a feature vector $\f_i \in \mathbb{R}^K$ of dimension $K$, $d_{i,j}$ can be computed as the \textit{Mahalanobis distance} \cite{mahalanobis1936} using a PSD \textit{metric matrix} $\M \succeq 0$:
\begin{align}
d_{i,j} = (\f_i - \f_j)^{\top} \M (\f_i - \f_j) .
\end{align}
$\M$ can be decomposed into $\M = \Q \Q^{\top}$ via \textit{Cholesky factorization} \cite{GoluVanl96}, where $\Q$ is a lower triangular matrix. 
In each neural layer, we first initialize an empirical covariance matrix $\E$ using available feature vectors $\{\f_i\}$.
We then apply Cholesky factorization to $\E^{-1}=\Q \Q^{\top}$.
Next, we designate a sparsity pattern in $\Q$ by setting to zero entries in $\Q$ whose amplitudes are small than factor $\zeta > 0$ times the average of the diagonals in $\Q$. 
Table\;\ref{tab:trainable_parameters} 
and Fig.\;\ref{fig:parameter_plot} show the number of trainable parameters for a $P$-layer unrolled network.
With a sparsity pattern set by a carefully chosen $\zeta$, the number of trainable parameters in $\Q$ in our network is $\mathcal{O}(K)$.


Using computed edge weights in \eqref{eq:edgeWeight}, one can compute a graph Laplacian matrix $\L_1 = \text{diag}(\W \1) - \W$, where $\1$ is the all-one vector.


\begin{table}[]
\begin{scriptsize}
\begin{center}
\caption{Trainable parameters for a $P$-layer network.
$M$ denotes the number of neurons in a dense layer.}
\label{tab:trainable_parameters}
\begin{tabular}{|c|c|c|}
\hline
\multirow{2}{*}{method} & \multicolumn{2}{c|}{trainable parameters} \\ \cline{2-3} 
 & type & count \\ \hline
MLP/CNN/GCN & weights/bias & $M[M(P-1)+P+K+2]+2$ \\ \hline
\multirow{2}{*}{SDR} & $Q$ & $\leq PK(K+1)/2$ \\ \cline{2-3} 
 & $Q,\lambda,\mu,\alpha_i$ & $\leq P(4+K(K+1)/2)$ \\ \hline
\end{tabular}
\end{center}
\end{scriptsize}
\end{table}

\subsection{Local Linear Embedding}

We compute a second graph Laplacian matrix $\L_2$ via \textit{local linear embedding} (LLE) \cite{Roweis2000NonlinearDR,ghojogh2020locally}. 
Specifically, we compute a \textit{sparse} coefficient matrix $\C \in \mathbb{R}^{N \times N}$, so that each feature vector $\f_i \in \mathbb{R}^K$ can be represented as a sparse linear combination of other feature vectors $\f_j$, $\forall j \,|\, j \neq i$. 
We first define matrix $\F \triangleq [\f_1; \ldots, ; \f_K] \in \mathbb{R}^{N \times K}$ that contains feature vector $\f_i$ as row $i$. 
We then formulate the following group sparsity problem:
\begin{align}
\min_{\C \in \cS^+} \left\| \F - \C \F \right\|^2_2 + \eta \| \C \|_{1,1}
\label{eq:groupSparse}
\end{align}
where $\cS^+$ is the set of symmetric matrices with zero diagonal terms and non-negative off-diagonal terms, and $\eta > 0$ is a parameter that induces sparsity in $\C$. 
Matrix symmetry and non-negativity are enforced, so that bi-directional positive edge weights can be easily deduced from $\C$. Using non-negative weights for LLE is called \textit{non-negative kernel regression} (NNK) in \cite{shekkizhar20icassp}.

The objective in \eqref{eq:groupSparse} contains two convex terms, where only the first term is differentiable. 
Thus, we optimize \eqref{eq:groupSparse} iteratively using \textit{proximal gradient} (PG) \cite{parikh13} given an initial matrix $\C$ that corresponds to the adjacency matrix of a k-nearest neighbor graph as input.
Specifically, at each iteration $t$, we first optimize the first term $\|\F - \C \F\|^2_2$ via gradient descent with step size $\delta$.
We then optimize the second term via soft-thresholding $T(c_{i,j})$: 
\begin{align}
\label{eq:solveLLE}
T(c_{i,j}) = \left\{ \begin{array}{ll}
c_{i,j} - \eta & \mbox{if}~~ c_{i,j} \geq \eta \\
0 & \mbox{o.w.}
\end{array} \right. .
\end{align}
$T(\cdot)$ combines the proximal operator for the $\ell_1$-norm and the projection operator onto $\cS^+$. 

As done in \cite{10.1007/978-981-13-0341-8_21,ghojogh2020locally}, the weights of the optimized $\C$ can be further adjusted using known labels in a semi-supervised manner:
the weights for the same-label (different-label) sample pairs are increased by parameter $\gamma >0$ (decreased by parameter $\mu >0$).

After $\C$ is obtained, we interpret it as an adjacency matrix and compute its corresponding graph Laplacian matrix $\L_2 = \text{diag}(\C \1) - \C$.
Finally, we compute a new graph Laplacian $\L$ as a \textit{conic combination} of $\L_1$ computed via feature distance specified by metric matrix $\M$ and $\L_2$ computed via LLE specified by coefficient matrix $\C$, \ie, 
\begin{align}
\L = \alpha_1 \L_1 + \alpha_2 \L_2, ~~~~~\alpha_i \geq 0 .
\label{eq:conicL}
\end{align}
$\alpha_i \geq 0$ ensures that the conically combined $\L$ is a Laplacian for a positive graph.
Trainable parameters for $\L$ consist of the two LLE adjustment parameters ($\gamma$ and $\mu$) and the two Laplacian weight parameters ($\alpha_1$ and $\alpha_2$).

\subsection{Loss Function and Inference}



As shown in Fig.\;\ref{fig:gdpaunroll}, 
we train parameters $\Q,\gamma,\mu,\alpha_1$ and $\alpha_2$
in each SDR layer in an end-to-end fashion via backpropagation \cite{Rumelhart:1986we}.
During training, a mean-squared-error (MSE) loss function is defined as

\vspace{-0.1in}
\begin{scriptsize}
\begin{align}
\begin{split}
\label{eq:loss_mse}
L= \bigg\Vert g\left[f_P\Big(f_{P-1}\big(\cdots f_1(\Q,\alpha_i,\gamma,\mu,\y,\z)\big)\Big)\right]-\hat{\x}_{\{M+1,...,N\}}\bigg\Vert_2^2,
\end{split}
\end{align}
\end{scriptsize}\noindent
where $g[\cdot] \triangleq \text{sign}\left\{\hat{x}_1v_1\v_{\{M+1,...,N\}}\right\}$ is the label prediction equation,
$\v$ is the first eigenvector of $\H$ computed by LOBPCG,
and $f_P(f_{P-1}(\cdots))$ are nested differentiable functions corresponding to the $P$-layers in our unrolled network. 
\eqref{eq:loss_mse} is essentially the MSE loss of entries $M+1$ to $N$ of $\x$ (unknown labels) compared to the ground-truth labels. 
We optimize the parameters using an off-the-shelf SGD optimizer.

During inference, test data is passed through the unrolled network, where the optimized $\Q,\gamma,\mu,\alpha_1$ and $\alpha_2$ are fixed.
$\Q$ is used to define the metric matrix $\M$ to construct $\L_1$.
$\gamma$ and $\mu$ are used to construct $\L_2$ together with the LLE weight matrix $\C$ learned from the test data via \eqref{eq:groupSparse}.
$\alpha_1$ and $\alpha_2$ are used to define $\L$. 
Finally, unknown labels are predicted.

\section{EXPERIMENTS}
\label{sec:results}
\subsection{Experimental Setup}
\label{ssec:experimentalsetup}

We implemented our unrolled network in PyTorch\footnote{results reproducible via code in \url{https://anonymous.4open.science/r/SDP_RUN-4D07/}.},
and evaluated it in terms of average classification error rate and inference runtime.
We compared our algorithm against the following six model-based schemes: 
i) a primal-dual interior-point solver that solves the SDP primal in Eq.~\eqref{eq:primal}, MOSEK, available in CVX with a Professional license \cite{cvx_link};
ii) a biconvex relaxation solver BCR \cite{shah16,bcr_link};
iii) a spectrahedron-based relaxation solver SDCut \cite{wang13,7431988,sdcut_link} that involves L-BFGS-B \cite{10.1145/279232.279236};
iv) an ADMM first-order operator-splitting solver CDCS \cite{8571259,cdcs_2020} with an LGPL-3.0 License \cite{cdcs_link} that solves the modified SDP dual in Eq.~\eqref{eq:SDP_dual2};
v) a graph Laplacian regularizer GLR \cite{pang17} with a box constraint $x_i \in [-1,1]$ for predicted labels;
and 
vi) baseline model-based version of our SDR network proposed in \cite{yang2021projectionfree} based on GDPA \cite{yang2021signed}.

In addition, we compared our network against four neural network schemes: 
vii) an unrolled 1-layer SDP classifier network that solves \eqref{eq:SDP_dual2} using a differentiable SDP solver in a Cvxpylayer library \cite{cvxpylayers2019,agrawal2020differentiating};
viii) a multi-layer perceptron (MLP) consisted of two dense layers;
ix) a convolutional neural network (CNN) consisted of two 1-D convolutional layers (each with a kernel size 1 and stride 2);
and
x) a graph convolutional network (GCN) \cite{Kipf:2016tc} consisted of two graph convolutional layers.
For GCN, the adjacency matrix is computed in the same way as the one used to compute the graph Laplacian in \eqref{eq:glr} and is fixed throughout the network training procedure.
For MLP, CNN and GCN, each (graph) convolutional layer is consisted of 32 neurons and is followed by group normalization \cite{Wu2018GroupN}, rectified linear units \cite{4082265} and dropout \cite{10.5555/2627435.2670313} with a rate 0.2.
A 1-D max-pooling operation with a kernel size 1 and stride 2 is placed before the dropout for CNN.
A cross entropy loss based on the log-softmax \cite{DBLP:journals/corr/BrebissonV15} is adopted for MLP, CNN and GCN.

We set the sparsity factor as $\zeta=0.9$ to initial lower triangular $\Q$,
the sparsity weight parameter as $\eta=0.01$ in \eqref{eq:solveLLE},
the Laplacian weight parameter in \eqref{eq:conicL} as $\alpha_1=\alpha_2=1$, 
and the LLE weight adjustment parameters as $\gamma=1$ and $\mu=1$. 
We set the convergence threshold of
i) LOBPCG to $10^{-4}$ with $200$ maximum iterations,
ii) the differentiable LP solver to $10^{-6}$ with $1000$ maximum iterations.
We set the learning rate for the SGD optmizer used in all methods to $10^{-2}$.
The maximum iterations for the optimization of $\Q$ and $\C$ was set to $1000$ for the pure model-based methods i, iii, iv, v and vi that involve graph construction.
For fast convergence, we set the convergence thresholds of CDCS and SDCut to $10^{-3}$, 
the maximum ADMM iterations in CDCS to $1000$,
the maximum iterations for L-BFGS-B in SDCut and the main loop in BCR to $100$,
and the Frobenius norm weight in SDCut to $100$. 
The number of epochs for the three data-driven networks, viii, ix and x, was set to $1000$. 
For the SDP unrolled network vii and our unrolled network it was $20$.
All computations were carried out on a Ubuntu 20.04.2 LTS PC with AMD RyzenThreadripper 3960X 24-core processor 3.80 GHz and 128GB of RAM.

We employed $17$ binary datasets freely available from UCI \cite{uci_link} and LibSVM \cite{libsvm_link}. 
For efficiency, we first performed a $K$-fold ($K\leq9$) split for each dataset with random seed 0, and then created 5 instances of 80\% training-20\% test split for each fold, with random seeds 1-5 \cite{10.5555/1671238}.
For the six model-based approach and the three data-driven networks, the ground-truth labels for the above 80\% training data were used for semi-supervised graph classifier learning \cite{yang2021projectionfree} and supervised network training.
For our SDR unrolled network, 
we further created a 75\% unroll-training-25\% unroll-test split for the 80\% training data, 
where, first, the ground-truth labels for the unroll-training data were used for the semi-supervised SDR network training together with the unroll-test data,
and second,
the learned parameters were used for label inference of the remaining 20\% test data.
The above setup resulted in sample sizes from 62 to 292.
We applied a \textit{standardization} data normalization scheme in \cite{classificationpami19} that first subtracts the mean and divides by the feature-wise standard deviation, and then normalizes to unit length sample-wise.
We added $10^{-12}$ noise 
to the dataset to avoid NaN's due to data normalization on small samples.

\subsection{Experimental Results}
\label{ssec:results}

\begin{table*}[]
\begin{small}
\begin{center}
\caption{Classification error rates (\%).
$K$ denotes feature count.}
\label{tab:error_rate}
\begin{tabular}{|c|c|c|c|c|c|c|c|c|c|c|c|c|c|}
\hline
\multirow{2}{*}{dataset} & \multirow{2}{*}{$K$} & \multicolumn{6}{c|}{model-based} & \multicolumn{6}{c|}{neural nets} \\ \cline{3-14} 
 &  & MOSEK & BCR & SDcut & CDCS & GLR & GDPA & MLP & CNN & GCN & \begin{tabular}[c]{@{}c@{}}SDR\\ $\T$\end{tabular} & \textbf{\begin{tabular}[c]{@{}c@{}}SDR\\ $\Q$\end{tabular}} & \textbf{\begin{tabular}[c]{@{}c@{}}SDR\\ $\Q$+LLE\end{tabular}} \\ \hline
australian & 14 & 20.14 & 15.65 & 15.65 & 15.65 & 16.67 & 15.51 & 17.39 & 17.83 & 19.57 & 18.70 & 15.65 & 16.95 \\ \hline
breast-cancer & 10 & 3.85 & 3.41 & 3.56 & 3.41 & 4.30 & 3.56 & 5.19 & 4.89 & 12.59 & 3.48 & 3.48 & 5.33 \\ \hline
diabetes & 8 & 35.16 & 32.94 & 31.76 & 31.63 & 33.59 & 35.03 & 32.31 & 36.15 & 33.08 & 30.98 & 30.00 & 29.62 \\ \hline
fourclass & 2 & 28.30 & 23.98 & 23.51 & 23.51 & 25.38 & 25.03 & 26.08 & 25.15 & 25.15 & 29.77 & 27.93 & 27.12 \\ \hline
german & 24 & 26.90 & 26.90 & 26.90 & 27.00 & 26.90 & 26.90 & 31.60 & 28.80 & 24.40 & 25.60 & 24.40 & 23.20 \\ \hline
haberman & 3 & 23.61 & 23.61 & 23.61 & 23.61 & 23.61 & 23.61 & 27.10 & 29.68 & 28.71 & 23.55 & 22.58 & 22.90 \\ \hline
heart & 13 & 20.37 & 18.89 & 18.89 & 18.89 & 18.52 & 18.89 & 24.81 & 24.07 & 23.70 & 17.41 & 18.89 & 21.11 \\ \hline
ILPD & 10 & 28.10 & 28.10 & 28.10 & 28.10 & 28.10 & 31.21 & 26.78 & 27.97 & 30.00 & 29.31 & 28.62 & 25.34 \\ \hline
liver-disorders & 5 & 30.00 & 27.86 & 30.71 & 30.00 & 29.29 & 30.71 & 37.86 & 39.29 & 44.29 & 41.33 & 36.00 & 34.67 \\ \hline
monk1 & 6 & 29.82 & 26.25 & 26.07 & 27.86 & 26.43 & 26.07 & 6.43 & 5.71 & 12.86 & 32.73 & 26.18 & 27.64 \\ \hline
pima & 8 & 35.16 & 32.68 & 31.90 & 32.03 & 33.59 & 36.47 & 33.08 & 32.69 & 35.00 & 31.37 & 28.08 & 29.62 \\ \hline
planning & 12 & 25.00 & 25.00 & 25.00 & 25.00 & 25.00 & 25.00 & 39.44 & 40.56 & 33.89 & 25.41 & 24.86 & 23.78 \\ \hline
voting & 16 & 11.40 & 10.70 & 10.70 & 12.09 & 11.40 & 10.70 & 3.95 & 2.79 & 10.47 & 10.93 & 3.72 & 4.19 \\ \hline
WDBC & 30 & 7.54 & 7.72 & 7.54 & 8.07 & 7.37 & 7.54 & 4.64 & 4.46 & 22.86 & 9.47 & 7.14 & 6.79 \\ \hline
sonar & 60 & 31.90 & 23.33 & 21.90 & 21.90 & 23.33 & 21.90 & 17.62 & 17.14 & 40.95 & 14.63 & 20.00 & 19.05 \\ \hline
madelon & 500 & 49.75 & 44.44 & 48.94 & 48.84 & 48.79 & 48.59 & 46.82 & 47.78 & 46.11 & 41.92 & 43.59 & 40.76 \\ \hline
colon-cancer & 2000 & 38.33 & 36.67 & 38.33 & 38.33 & 38.33 & 38.33 & 28.33 & 26.67 & 38.33 & 32.31 & 28.33 & 23.08 \\ \hline
\textbf{avg.} & $-$ & 26.20 & 24.01 & 24.30 & 24.47 & 24.74 & 25.00 & 24.08 & 24.21 & 28.35 & 24.64 & \textbf{22.91} & \textbf{22.42} \\ \hline
\end{tabular}
\end{center}
\end{small}
\end{table*}


\begin{figure}
\begin{center}
\includegraphics[trim={0cm 0cm 0cm 0cm},clip,width=0.95\linewidth]{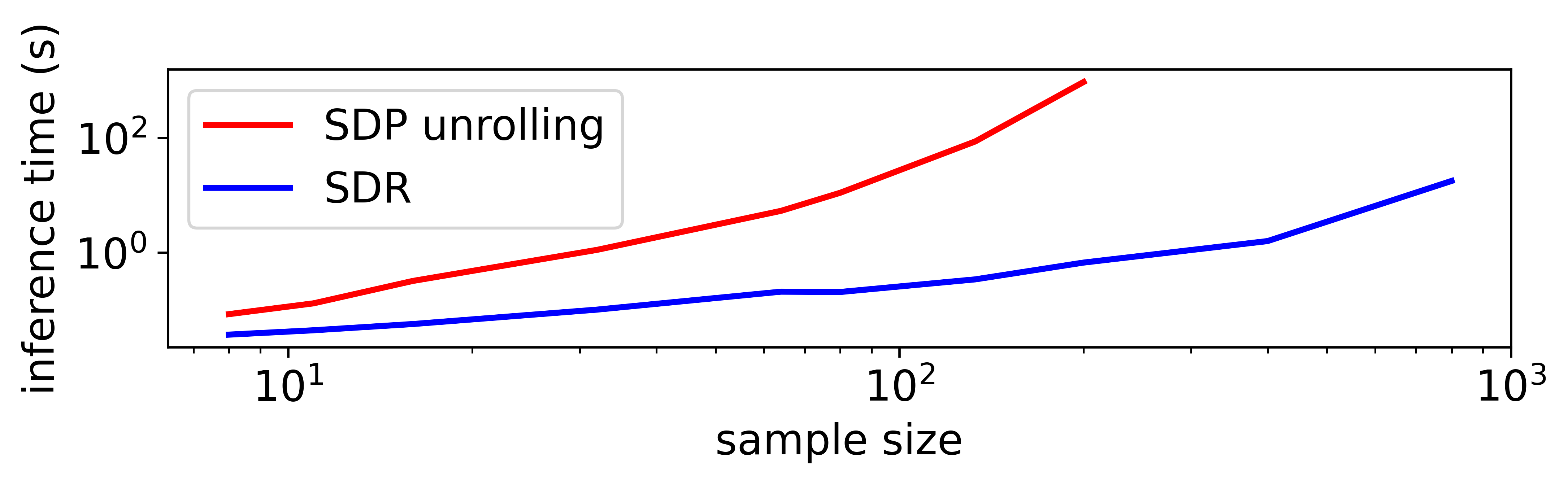}
\end{center}
\vspace{-0.25in}
\caption{Inference time of SDR v.s. naive SDP unrolling.
Our speedup is over $1400\times$ when the sample size is 200.}
\label{fig:time_plot}
\end{figure}

We first show in Fig.~\ref{fig:time_plot} the inference runtime of our SDR network compared to a SDP unrolled network that na\"{i}vely unrolls the PSD cone projection using the same Cvxpylayer python library described earlier. 
It is clear that our SDR network is substantially faster in inference than the na\"{i}ve SDP unrolled network, with a speedup that is over $1400\times$ when the sample size is $200$.

We next show in Table~\ref{tab:error_rate} the classification error rates of the six model-based schemes, 
namely MOSEK,
BCR,
SDcut,
CDCS,
GLR
and GDPA,
the three data-driven networks,
namely MLP,
CNN 
and GCN,
and the three variants of our SDR network where a single SDR layer optimizes
i) a $K\times J$ matrix $\T$ for $\M=\T\T^\top$ where $J=2$ is the pre-defined rank,
ii) our proposed lower-triangular matrix $\Q$ sparsified by $\zeta$,
and
iii) $\Q$ plus LLE weight adjustment parameters $\gamma$ and $\mu$ and Laplacian weighting parameters $\alpha_1$ and $\alpha_2$.

We first observe that, in general, with an appropriate choice of the trainable parameters, SDR $\Q$ and SDR $\Q$+LLE outperformed on average all model-based schemes and were competitive with data-driven schemes.
SDR $\Q$+LLE on average performed better than SDR $\Q$ thanks to
the four additional trainable parameters $\gamma$, $\mu$, $\alpha_1$ and $\alpha_2$.

We observe also that the three data-driven schemes, MLP, CNN and GCN, performed on average slightly worse than SDR $\Q$ and SDR $\Q$+LLE.
This can be explained by the relatively large number of trainable parameters that may cause overfitting.
For example, MLP is consisted of 1602 trainable parameters with 32 neurons in each of the two dense layers during training on the dataset \texttt{australian},
while 1-layer SDR $\Q$ and SDR $\Q$+LLE have at most 105 and 109 trainable parameters, respectively.
We see also that SDR $\T$, SDR $\Q$ and SDR $\Q$+LLE learned faster than the three data-driven schemes with only 20 epochs in training stage compared to 1000 epochs for MLP, CNN and GCN.
We note further that our unrolled network is by design more interpretable than the three generic black box data-driven implementations, where each neural layer is an iteration of an iterative algorithm.

We observe that SDR $\Q$ and SDR $\Q$+LLE outperformed SDR $\T$, demonstrating that our proposed parameterization of graph edge weights at each neural layer is better than simple low-rank factorization $\M = \T \T^{\top}$.
For SDR $\T$, SDR $\Q$ and SDR $\Q$+LLE, the noticeably worse performance on the dataset \texttt{liver-disorders} compared to the model-based schemes may be explained by the fact that the optimizer was stuck at a bad local minimum. 

\begin{table}[]
\begin{center}
\caption{Classification error rates (\%) on the dataset \texttt{sonar} using our SDR $\Q$+LLE with $P$ layers.}
\label{tab:SDR_LLE}
\begin{tabular}{|c|c|c|c|}
\hline
$P$ & 1 & 2 & 3 \\ \hline
error rate (\%) & 19.05 & 16.59 & 16.10 \\ \hline
\end{tabular}
\end{center}
\end{table}

We show in Table\;\ref{tab:SDR_LLE} the classification error rate of our SDR $\Q$+LLE with 1, 2 and 3 SDR layers on the dataset \texttt{sonar}. 
We see that as the number of layers $P$ increases, the classification error rates are reduced at the cost of introducing more network parameters.
This indicates that our SDR network is resilient to overfitting when the number of trainable parameters increases by a factor of $P$.

\section{CONCLUSION}
\label{sec:conclude}
To facilitate algorithm unfolding of a proximal splitting algorithm that requires PSD cone projection, using binary graph classifier as an illustrative example, we propose an unrolling strategy via GDPA linearization.
Specifically, we replace the PSD cone constraint in the semi-definite programming relaxation (SDR) of the classifier problem by ``tight possible'' linear constraints per iteration, so that each iteration requires only computing a linear program (LP) and the first eigenvector of the previous matrix solution. 
After unrolling iterations of the projection-free algorithm into neural layers, we optimize parameters that determine graph edge weights in each layer via stochastic gradient descent (SGD).
Experiments show that our unrolled network outperformed pure model-based classifiers, and had comparable performance as pure data-driven schemes while employing far fewer parameters.


\bibliographystyle{aaai}
\bibliography{ref2}

\end{document}